# DSNet: An Efficient CNN for Road Scene Segmentation

Ping-Rong Chen[1]   Hsueh-Ming Hang[1]
[1]National Chiao Tung University
{james50120.ee05g, hmhang}@nctu.edu.tw

Sheng-Wei Chan[2]   Jing-Jhih Lin[2]
[2]Industrial Technology Research Institute
{ShengWeiChan, jeromelin}@itri.org.tw

## Abstract

*Road scene understanding is a critical component in an autonomous driving system. Although the deep learning-based road scene segmentation can achieve very high accuracy, its complexity is also very high for developing real-time applications. It is challenging to design a neural net with high accuracy and low computational complexity. To address this issue, we investigate the advantages and disadvantages of several popular CNN architectures in terms of speed, storage and segmentation accuracy. We start from the Fully Convolutional Network (FCN) with VGG, and then we study ResNet and DenseNet. Through detailed experiments, we pick up the favorable components from the existing architectures and at the end, we construct a light-weight network architecture based on the DenseNet. Our proposed network, called DSNet, demonstrates a real-time testing (inferencing) ability (on the popular GPU platform) and it maintains an accuracy comparable with most previous systems. We test our system on several datasets including the challenging Cityscapes dataset (resolution of 1024 ×512) with an mIoU of about 69.1 % and runtime of 0.0147 second per image on a single GTX 1080Ti. We also design a more accurate model but at the price of a slower speed, which has an mIoU of about 72.6 % on the CamVid dataset.*

## 1. Introduction

With the fast development of automated driving systems, a stable and reliable surrounding scene analysis becomes essential for a safe driving environment. The deep-learning based image semantic segmentation is one of the best solutions because it is sufficiently robust in analyzing the complicated environments. It partitions a captured image into several regions and recognizes the class (object) of every pixel, so it can be viewed as pixel-level classification. Different from image classification, the image semantic segmentation identifies the object classes in images and also finds the locations of objects in images. In addition, it provides precise object boundary information. Nevertheless, the high accuracy of image semantic segmentation is often at the high complexity of a CNN model without consideration of inference time, resulting in a difficult implementation on several light devices. Therefore, a fast and efficient CNN model is very desirable and imperative for a practical semantic segmentation system.

Recently, an encoder-decoder architecture is popular for semantic segmentation. The encoder is usually a classification network, such as VGG [28], ResNet [10], and DenseNet [12]. It employs a series of down-sampling layers to condense the information. However, the down-sampling operation drastically reduces the detailed spatial information which is quite important for the image semantic segmentation task. To address this issue, some decoders are designed to recover the spatial resolution by using the up-sampling process. Deconvolution is commonly used to produce a learnable up-sampling process in many popular networks, such as DeconvNet [22] and FCN [19]. Un-pooling used in SegNet [1] is another method to up-sample the feature maps by reusing the max-pooling indices produced by the encoder. On the other hand, some networks are constructed without a decoder network but retain the detailed spatial information from the encoder part, such as DeepLab v2 [4], DeepLab v3 [5], and PSPNet [35]. They remove some down-sampling layers and apply the dilated convolution, which can maintain the spatial resolution without sacrificing the receptive field. Although this method can improve accuracy, the enlarged feature maps often significantly slow down the processing, especially for a deep architecture together with large feature maps. Also, DeepLab v3+ [6] includes a decoder network to combine the multi-scale information to obtain better results. These previous works give us clues in constructing a fast network that is able to capture multi-scale information without using dilated convolution.

Recently, some efficient semantic segmentation networks have been proposed, such as ENet [23] and ERFNet [26]. ENet is constructed based on the concept of SegNet, but it is much slander than the latter and thus offers a light and fast architecture. Moreover, ENet uses dilated convolution and stacked residual layers to deepen the network so that the accuracy can be maintained. ERFNet is a wider version of ENet and uses deconvolutional layers for the up-sampling process. Also, they adopt the factorized filter structure [30] to separate a two-dimensional filter into two one-dimensional filters in the convolutional layer and thus considerably reduce the number of parameters. Further, both of ENet and ERFNet use an early down-sampling scheme and an extremely narrow architecture compared to the schemes with a heavy encoder such as VGG16 and ResNet101 (top-ranked on ILSVRC [27]). In this work, we adopt a similar idea in constructing a narrow architecture, but DenseNet is selected as the backbone model instead of ResNet because it combines the multi-scale information more frequently, which may be more appropriate for semantic segmentation purpose.

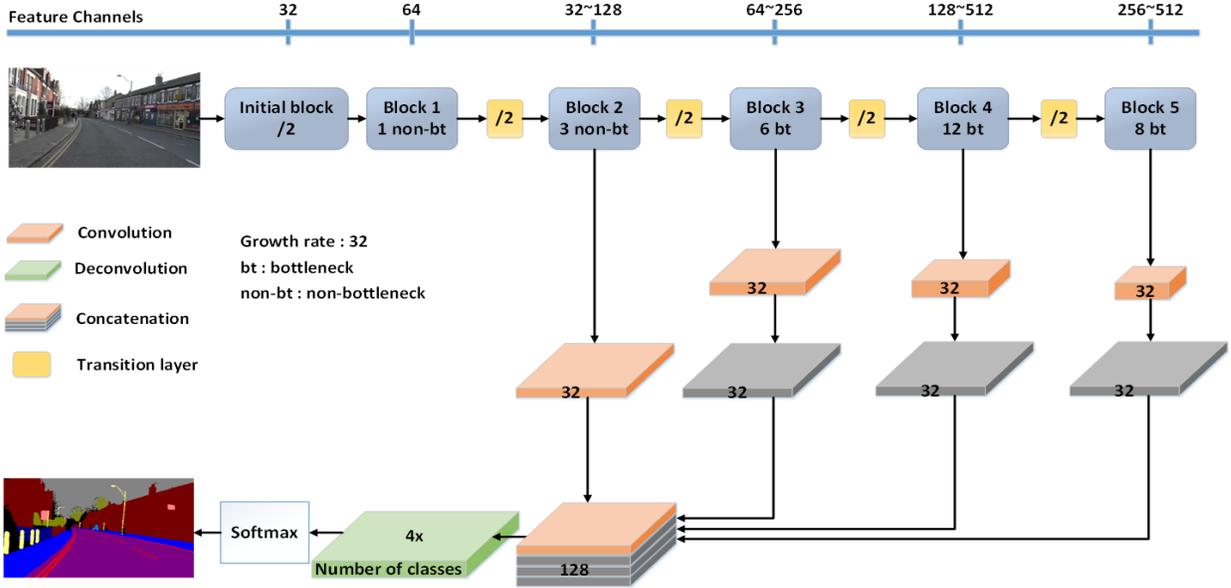

Figure 1: The architecture of fast dense segmentation network (DSNet-fast). The encoder is constructed based on the deep dense unit described in Figure 2. The decoder is designed as light as possible. The accurate version (DSNet-accurate) removes the down-sampling operation in the initial block. Additionally, the convolutional layer connected to Block 2 is removed, performing 96 channels at the concatenation layer rather than 128 channels in DSNet-fast.

## 2. Proposed Network

In this paper, the target is to construct a fast network architecture without degrading its accuracy. Up to now, constructing a CNN is often empirical, and it is hard to predict in advance the results of a modified network. Our aim is to design a favorable architecture for real-time applications. We conducted a series of experiments as to be described in section 3. Here, we first give an overview of the proposed network, Dense Segmentation Network, DSNet, in brief. The entire architecture is shown in Figure 1. Mainly, the network architecture includes two parts, the encoder and the decoder. The details are described as follows.

### 2.1. Encoder

The encoder part is constructed based on the concept of DenseNet because it achieves high performance with narrow layers, resulting less overall computational cost. The encoder consists of one initial block, 4 non-bottleneck units (without 1×1 convolution), and 26 bottleneck units (with 1×1 convolution). The early down-sampling operation (convolution with a stride of 2) is employed at the initial block to shrink the size of feature maps and to speed up the network. Meanwhile, the output channel of the initial block is set to 32, the growth rate is set to 32, which represents how many feature maps are generated in one dense unit, and the channels are compressed with a ratio of 0.5 in the transition layer before a pooling operation to reduce the complexity. The bottleneck units adopt a 1×1 convolution to reduce the number of channels. Because the number of channels is quite small in Block 1 and Block 2, it is not necessary to decease the channel numbers there. Thus, the non-bottleneck units are adopted in Block 1 and Block 2 rather than the bottleneck units.

### 2.2. Decoder

In our experiments, we find that a decoder with heavy structure does not seem to provide much accuracy improvement. Hence, simplifying the decoder is a feasible way to speed up the network. So, we reduce the number of channels to 32 by employing four convolutional layers after the Block 2, Block 3, Block 4, and Block 5. Furthermore, all the feature maps are up-sampled to the resolution of Block 2 to concatenate them together. In the end, a deconvolutional layer with a 4 up-sampling rate is used to recover the spatial resolution for the final dense segmentation. By using this simple decoder, the computational complexity is considerably decreased and the accuracy can be maintained at the same time.

### 2.3. Details of Encoder

In this paper, we design two modifications on the original dense unit to make a trade-off between the accuracy and the speed. First, we modify the composition unit from BN-ReLU-Conv [11] to Conv-BN-ReLU. Even though the full pre-activation unit (BN-ReLU-Conv) is claimed to improve the results, it is not possible to merge BN layers into Conv layers (as discussed in [13]) during the testing phase. So, for speed consideration, we redesign the DenseNet architecture by using the Conv-BN-ReLU units to replace the BN-ReLU-Conv units.

Second, we slightly modify the dense unit by inserting another convolutional layer at the end of bottleneck and the non-bottleneck architectures, as shown in Figure 2. By adopting this modification, the network can be deepened at a low computational cost because the preceding layer is sufficiently narrow. Also, the additional convolutional layer can enlarge the receptive field to capture large-scale objects and to produce a better result on the high resolution dataset, such as Cityscapes dataset.

### 2.4. Accurate Version

In our experiments, we find that the convolutional layers operated on the large feature maps is important for an accurate pixel-level classification, especially for processing a low resolution image. Thus, we propose another architecture to deal with the low resolution images. This architecture removes the early down-sampling operation in the initial block in Figure 1 so that the larger feature maps are fed to the rest of the network.

Meanwhile, in the decoder, one convolutional layer connected to Block 2 is removed and all the feature maps are up-sampled to the resolution of Block 3. So, the number of channels after the concatenated layer becomes 96 and the last deconvolutional layer up-samples the feature maps by a factor of 4. This architecture is called DSNet-accurate (accurate dense segmentation network). In summary, we proposed two architectures to deal with different input resolutions. We name the architecture with the early down-sampling layer as *DSNet-fast*. And, the architecture without the early down-sampling operation is called as *DSNet-accurate*.

## 3. Experiments

As mentioned earlier, a series of experiments are carried out to show the pros/cons of various structure components. The results give us the clues in designing DSNet. We now explain the experiments in details in this section. DSNet came from the architecture of FCN. FCN is an encoder-decoder network and its decoder combines the multi-scale information in order to produce an accurate prediction. Portions of our experiments are implemented on the modified FCN, which increases the number of channels in the decoder compared to the original FCN. The modified FCN is shown in Figure 3. Mainly, we divide the experiments into two parts, Encoder Experiments and Decoder Experiments. Before describing them, we first describe the dataset used to benchmark the performance and the parameter setting in training the networks.

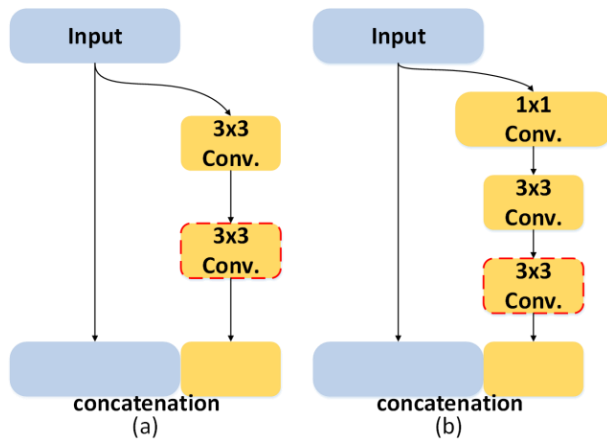

Figure 2: The modified deep dense units by inserting another convolutional layer (red dotted block). (a) Non-bottleneck architecture. (b) Bottleneck architecture.

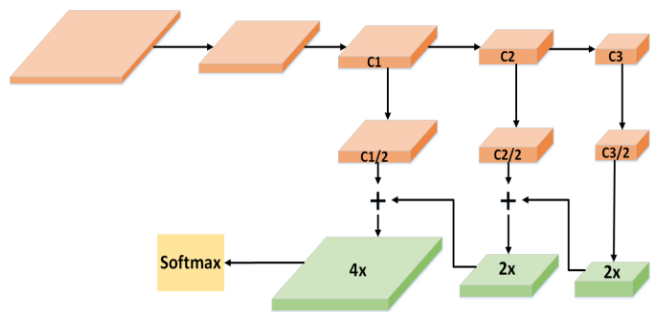

Figure 3: The modified FCN. C1, C2, and C3 denote the number of channels. The decoder reduces the channels to half and recover the original spatial resolution.

### 3.1. Dataset

We use two popular road-scene datasets to evaluate all the networks in this paper. The first one is CamVid [2] consisting of 367 training images, 101 validation images and 233 testing images. The resolution of all images is 480×360 and there are in total 11 classes in the dataset. The CamVid dataset can show the importance of detailed information because of the low resolution images.

The second dataset is Cityscapes [7], which is a larger dataset for semantic understanding of urban street scenes. All images are at 2048×1024 resolution and there are 19 classes for training. Two kinds of annotations are provided, fine-annotation and coarse-annotation. In this paper, we only use the fine-annotated dataset to train and evaluate the networks. It is composed of 2950 training images, 500 validation images, and 1525 test images. For speed consideration, the images are down-sampled by a factor of 2 (horizontally and vertically) in some experiments. The Cityscapes dataset can show the influence of receptive field on the networks when the input images are of high resolutions.

### 3.2. Implementation Details

All the experiments are conducted on the Pytorch framework [24] with a single Maxwell Titan X GPU. The optimizer is Stochastic Gradient Descend (SGD), with a weight decay of 0.0005, a momentum of 0.9, a batch size of 4 and a base learning rate of 0.05. Inspired by [4, 35], the learning rate is adjusted after every iteration according to the following equation:

$$lr = lr_{base} \times \left(1 - \frac{iteration}{total\ iterations}\right)^{power} \quad (1)$$

with a power of 0.9. Also, a total of 13800 iterations (150 epochs) is set in the training on the CamVid dataset, and 74400 iterations (100 epochs) is set in the training on the Cityscapes dataset.

In addition, we adopt a class balancing strategy to compensate small-size classes. Inspired by [23, 26], the class weightings are calculated by the equation (2):

$$w_c = \frac{1}{log(p_c+k)} \quad (2)$$

where $k$ is a constant set to 1.1 and $p_c$ represents the probability of the presence of class $c$ in pixel-level; then, these class weightings are divided by the maximum to normalize their values into [0, 1], so that the other hyper-parameters can be fixed without affecting the convergence in training.

### 3.3. Encoder Experiments – Ablation Study

In general, the encoder network for semantic segmentation is strongly related to the image classification network. Moreover, the residual unit has been proved that it can significantly improve the accuracy if the depth of the CNN is deeper. Hence, we modify the encoder of FCN by replacing VGG16 by ResNet50 to construct the FCN-ResNet50 network. However, according to Table 1, we find that the performance varies on different datasets. The performance of FCN-ResNet50 is worse than that of FCN-VGG16 on the CamVid dataset, but the accuracy of FCN-ResNet50 is significantly improved on the Cityscapes dataset. We suppose that this is due to different input resolutions. Thus, we investigate the impact of the input resolution by using the following experiments.

| Method | Dataset | mIoU (%) | Frame Rate (FPS) |
|---|---|---|---|
| FCN-VGG16 | CamVid | 67.5 | 39.8 |
|  | Cityscapes | 65.2 | 4.1 |
| FCN-ResNet50 | CamVid | 65.0 | 38.1 |
|  | Cityscapes | 67.4 | 5.1 |

Table 1: Results of FCN-VGG16 and FCN-ResNet50 on CamVid test set (480×360) and on Cityscapes validation set (2048×1024)

| Method | mIoU (%) | Frame Rate (FPS) | Model Size (MB) |
|---|---|---|---|
| FCN-VGG16 | 56.9 | 39.8 | 72.6 |
| FCN-VGG-ED | 52.2 | 60.6 | 71.1 |

Table 2: Results of FCN-VGG16 and FCN-VGG-ED on CamVid test set (training from scratch).

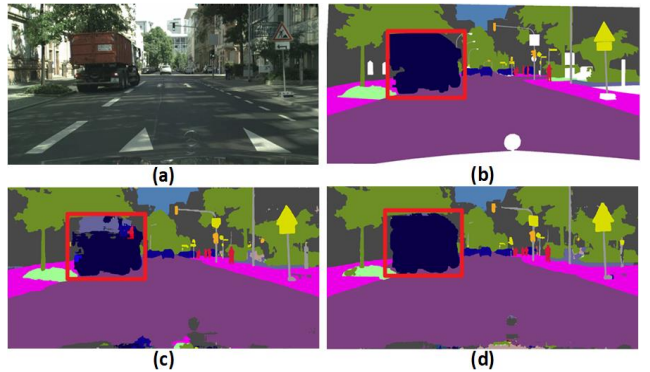

Figure 4: An output sample that shows the importance of receptive field. (a) Input image (b) Ground truth (c) FCN-VGG16 output (d) FCN-ResNet50 output.

First, observing the segmentation maps in Figure 4, we find that the truck object estimated by FCN-VGG16 is fragmented and incomplete. In contrast, FCN-ResNet50 is able to capture the truck more accurately. One explanation is that the receptive field of ResNet50 is larger than that of VGG16; that is, the ResNet50 is able to recognize large size objects and thus results in a more accurate estimation on the high resolution images.

Next, we study the performance on the low resolution images. We investigate the problem based on the structure of VGG16 and ResNet50. In the first few layers, the structure of ResNet50 has an early down-sampling layer, and then followed by a down-sampling layer (max-pooling layer). Heuristically, these consecutive down-sampling operations may reduce the feature maps to a too small size.

Thus, it is difficult to retain the detailed spatial information and thus leads to a poor segmentation result. In order to verify this speculation, we replace the first four convolutional layers and two max-pooling layers of VGG16 by one early down-sampling layer followed by a down-sampling layer to construct FCN-VGG-ED.

Because there is no pre-trained model for the modified network, FCN-VGG16 and FCN-VGG-ED (Early Down-sampling) in Table 2 are trained with the random initialization (training from scratch). The results show that the FCN-VGG-ED performance degrades considerably and we thus confirm that the early down-sampling layer is one of the reasons causing FCN-ResNet50 performance degradation on low resolution images. On the other hand, the number of parameters may be another reason causing the degradation in FCN-VGG-ED. Further investigation is explained in section 4.1 using Table 5 and Table 6. In summary, we have the following observations based on the results of encoder experiments.

- In order to retain the detailed spatial information, the convolution operation performed on the large feature maps is important for semantic segmentation.
- In order to capture the long-range information and large-scale objects, a deep architecture is needed for high resolution images.
- The early down-sampling layer speeds up the operation significantly but it damages the small-size image results.

In brief, we need a deep architecture to process a high resolution image. Also, we hope the network is capable of adding extra convolutional layers for large-size feature maps but we also want to have a low computational cost. Thus, the trade-off between using the early down-sampling layer and the extra convolutional layers is a critical issue in designing a fast neural network. The above observations give us clues in designing the encoder part in Figure 1. Because of the low complexity in every dense unit, we can insert additional convolutional layer to process the large feature maps (for example, Block 1 in Figure 1) and thus more detailed spatial information can be retained.

### 3.4. Decoder Experiments

After designing the encoder network, in order to train an initial encoder model, we add one global average pooling layer and one fully connected layer to convert the encoder into a classification network. Then, the network is pre-trained on a large dataset, ImageNet, to generate a good initial encoder model, which seems to be a good starting point for the complete system, encoder plus decoder although it is trained on the classification dataset (not segmentation dataset). Then, the following experiments on the decoder are conducted with the pre-trained encoder.

We first discard the attached pooling layer and the fully connected layer of the encoder, and connect it to the FCN-based decoder, fusing the feature maps by the summation operation. Further, inspired by DenseNet, we adopt the concatenation operation to fuse the feature maps in place of the summation operation in the decoder. However, concatenating the feature maps directly increases the number of channels (wide decoder) and results in a high computational cost in the following layers. Therefore, we reduce the output channels to half at every convolutional layer at the decoder to reduce the complexity (narrow decoder). According to the results in Table 3, there is no obvious difference among the accuracy of these three architectures. But the model with narrow decoder speeds up the network using fewer parameters. Hence, simplifying the decoder seems to be a way to construct an efficient network.

| Decoder | mIoU (%) | Frame Rate (FPS) | Model Size (MB) |
|---|---|---|---|
| Summation | 68.8 | 45.1 | 27.9 |
| Concat-wide | 68.7 | 41.4 | 30.9 |
| Concat-narrow | 68.7 | 50.6 | 18.0 |

Table 3: Fusion methods in the decoder. Testing on the Cityscapes validation set at 1024×512 resolution.

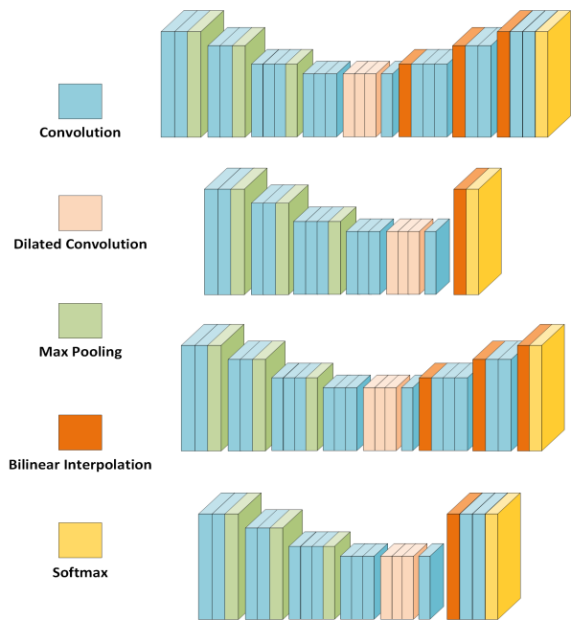

Figure 5: Variations of Decoder. From top to bottom: (a) Model-1, (b) Model-2, (c) Model-3, (d) Model-4

| Method | mIoU (%) | Frame Rate (FPS) | Model Size (MB) |
|---|---|---|---|
| Model-1 | 64.1 | 24.3 | 70.6 |
| Model-2 | 63.0 | 34.5 | 59.1 |
| Model-3 | 63.9 | 27.0 | 70.2 |
| Model-4 | 64.2 | 30.3 | 60.3 |

Table 4: Results of four decoders on CamVid test set.

For the purpose of designing a light decoder, we conduct variations of the decoder, as Figure 5 shows. All of them have the identical encoder, which is similar to the front-end module in Dilation 8 [33]. The 14th convolutional layer is inserted to adjust the number of channels for the decoder. In total, four decoders are designed and tested. Model-1 adopts the SegNet-like decoder but replaces the un-pooling layers by the bilinear interpolation layers. Model-2 discards the decoder network and up-samples the feature maps directly to the input resolution without additional convolutional layers. It can be viewed as an architecture without the decoder. Model-3 recovers the spatial resolution gradually but it removes the last two convolutional layers compared to Model-1. Model-4 up-samples the feature maps directly to the input resolution and uses two convolutional layers to recover the detailed information.

Table 4 shows the results of four architectures in Figure 5 when the CamVid dataset is tested. The results of these 4 models are very close, which confirms that the decoder plays a lesser role in improving the overall performance. Thus, we can simplify the decoder to speed up the network. Additionally, the results show that a good decoder can slightly improve the accuracy, so we design the network with a moderate decoder for accuracy consideration. Moreover, among Model-1, Model-3, and Model-4, Model-4 is fast and using fewer parameters, and thus is preferred for constructing an efficient network. Thus, the decoder of Model-4 is a good choice for our decoder.

In summary, the decoder experiments give us some observation as follows.

- Using a narrow decoder is able to speed up the network and it provides similar accuracy results compared to a wide decoder.
- Up-sampling the feature maps to a large size and/or using additional convolutional layers to recover the information can produce more accurate results.

Therefore, we remove the decoder of FCN-DenseNet and use the narrow convolutional layers followed by the bilinear interpolation layers to produce the large-size feature maps. A concatenated layer is employed to combine the feature maps, and a deconvolutional layer is used to recover the detailed information, to fuse the feature maps and to determine the final estimation. This architecture is our proposed DSNet.

## 4. Performance of DSNet

As mentioned above, two networks are proposed in this paper, DSNet-fast (Figure 1) and DSNet-accurate. These two architectures can be used to process different resolution images. In this section, we report the experimental results of the proposed network on CamVid and Cityscapes datasets. Also, we compare them with the other state-of-the-art networks to examine the effectiveness of the proposed method.

| Method | mIoU (%) | Global Acc. (%) |
|---|---|---|
| DeepLab-LFOV [4] | 61.6 | - |
| Bayesian SegNet [16] | 63.1 | 86.9 |
| Dilation8 [33] | 65.3 | 79.0 |
| EDANet [18] | 66.4 | 90.8 |
| FC-DenseNet103 [15] | 66.9 | 91.5 |
| ICNet [34] | 67.1 | - |
| G-FRNet [14] | 68.0 | - |
| DCDN [8] | 68.4 | 91.4 |
| SDN [9] | 71.8 | 92.7 |
| DSNet-fast | 68.6 | 91.7 |
| DSNet-accurate | 72.6 | 92.7 |

Table 5: Comparison of DSNet and other schemes on CamVid test set.

| Method | Frame Rate (FPS) | Model Size (MB) |
|---|---|---|
| DSNet-fast | 81.9 | 11.9 |
| DSNet-accurate | 58.2 | 11.6 |

Table 6: The speed of DSNet running on 480×360 resolution with 11 categories (CamVid dataset).

### 4.1. Results on CamVid

In this subsection, the CamVid dataset is used to evaluate the performance of DSNet. In addition to DSNet-fast, the DSNet-accurate is employed to process the small resolution images. Here, both DSNet-fast and DSNet-accurate adopt the pre-trained encoder on ImageNet. After pre-training, the data augmentation strategy (horizontal flip and pixel translation) is employed to produce the robust prediction. Also, we find that decaying the learning rate by equation (1) can slightly improve the accuracy.

The results are shown in Table 5. DSNet-accurate sacrifices the inference speed but its accuracy is higher than DSNet-fast for about 4 % mIoU. In addition, Table 6 shows that the number of parameters in DSNet-accurate is less than DSNet-fast due to the elimination of a convolution layer in the decoder. Here, we already know that the parameters used in the decoder provide less effective for accuracy improvement. Also, the number of the parameters used in first few layers is identical in both DSNet-fast and DSNet-accurate, which indicates the size of feature maps is supposed to be more important than using extra parameters. Thus, the results in Table 5 and Table 6 are consistent with our conjecture in section 3.3 that the feature map size has a significant influence on the accuracy in semantic segmentation.

On the other hand, compared to the other state-of-the-art methods, DSNet-accurate shows an outstanding performance in processing the low resolution images (480×360). Furthermore, according to the experimental results in Figure 6, we find that DSNet-accurate is indeed capable of retaining more detailed information and capturing the small objects compared to DSNet-fast. Table 6 lists the frame rate of DSNet-fast and DSNet-accurate. Both of them process a 480×360 RGB image for more than 55 frames/sec, which demonstrates the real-time testing (inferencing) ability. Thus, if the computing device is sufficiently powerful or the input size is small, the DSNet-accurate is also an appropriate architecture to balance speed and accuracy for the real-time applications.

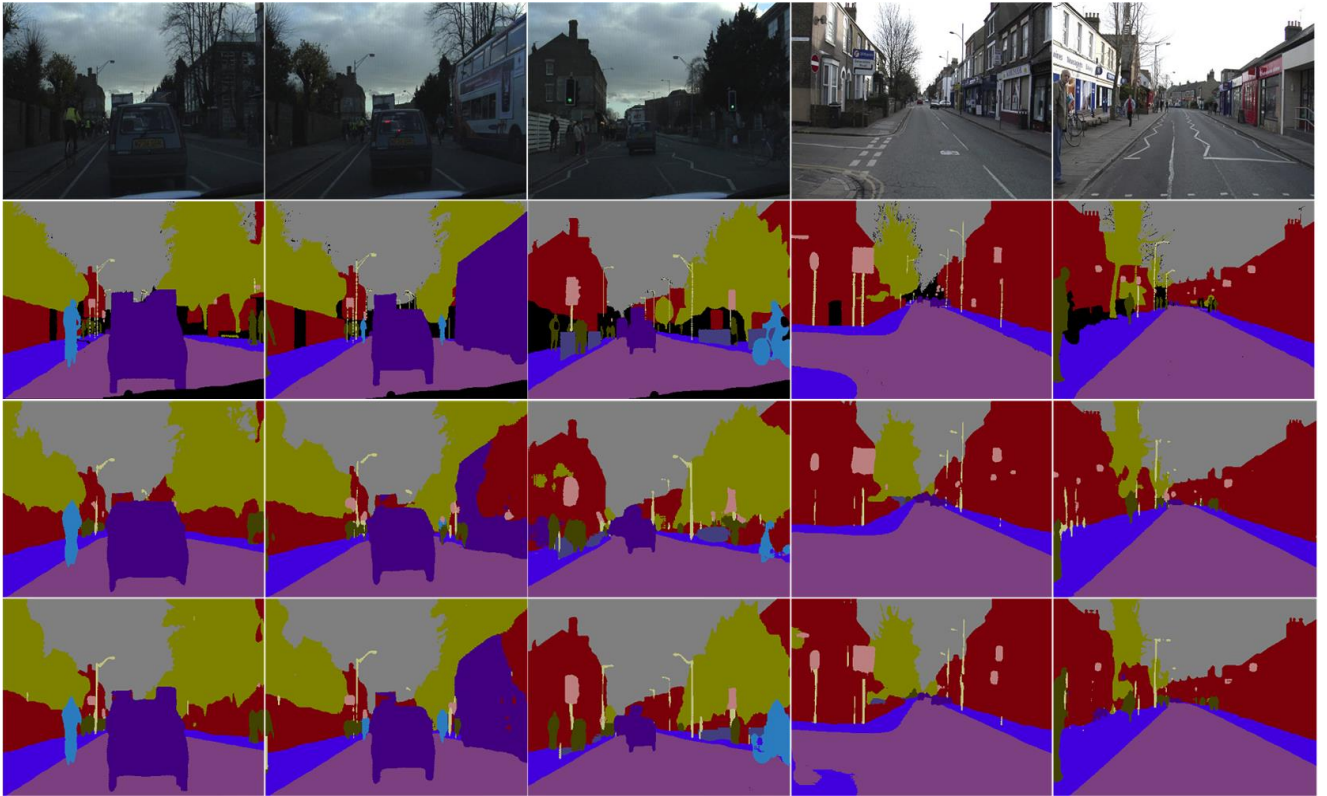

Figure 6: The results of DSNet on CamVid test set. From top to bottom: (a) Input image, (b) Ground truth, (c) DSNet-fast output, (d) DSNet-accurate output

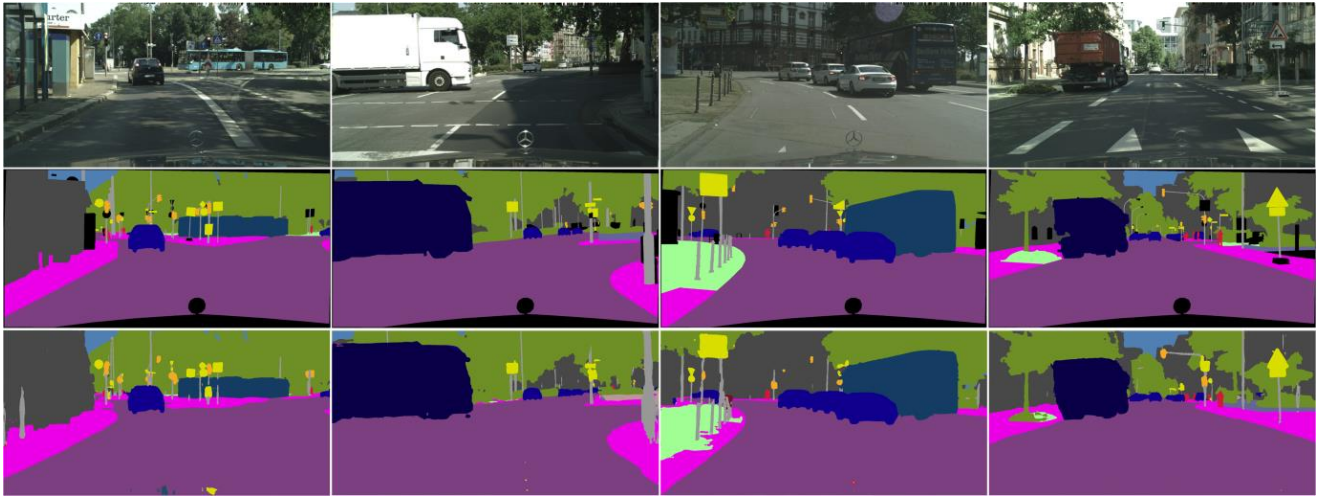

Figure 7: Results of DSNet on Cityscapes validation set. From top to bottom: (a) Input image, (b) Ground truth, (c) DSNet-fast output

### 4.2. Results on Cityscapes

We also tested our systems on the Cityscapes dataset. For the speed consideration, only DSNet-fast is benchmarked on this high resolution dataset. At the training step, we resize the image and its ground truth (map) to 1024×512 in order to speed up the network training. During inference, the input image is at 1024×512 resolution, but the output segmentation maps are up-sampled to the full resolution (2048×1024) corresponding to their ground truths for evaluation. In addition to decaying the learning rate by equation (1), we find that adjusting the weight decay from 0.0005 to 0.0001 and adopting the dropout strategy [29] at the end of every dense unit with a drop rate of 0.1 can further improve the accuracy for DSNet-fast. Also, the pre-trained encoder and the data augmentation strategy are used in training to strengthen the feature representation. The best model of DSNet-fast can achieve 71.5% mIoU on the validation set. Some output samples are displayed in Figure 7.

| High Accuracy Network | | | | |
|---|---|---|---|---|
| Method | Cityscapes data | Additional data | mIoU (%) | Runtime (s) |
| DRN_CRL_Coarse [37] | Fine, Coarse | ImagNet | 82.8 | - |
| DPC [3] | Fine, Coarse | ImageNet, COCO [17] | 82.7 | - |
| RelationNet_Coarse [36] | Fine, Coarse | ImageNet | 82.4 | - |
| SSMA [32] | Fine, Coarse, Stereo | ImageNet | 82.3 | - |
| High Speed Network | | | | | |
|---|---|---|---|---|---|
| Method | Cityscapes data | Additional data | mIoU (%) | Runtime (s) | |
| | | | | Titan X | 1080 Ti |
| SegNet [1] | Fine | ImageNet | 56.1 | 0.0600 | - |
| ENet [23] | Fine | - | 58.3 | 0.0130 | - |
| SQ [31] | Fine | ImageNet | 59.8 | 0.0600 | - |
| ESPNet [20] | Fine | - | 60.3 | 0.0089 | - |
| ESPNetv2 [21] | Fine | ImageNet | 62.1 | 0.0120 | - |
| ContextNet [25] | Fine | - | 66.1 | 0.0238 | - |
| EDANet [18] | Fine | - | 66.3 | 0.0123 | 0.0092 |
| EDANet [18] | Fine, Validation | - | 67.3 | 0.0123 | 0.0092 |
| ERFNet [26] | Fine | - | 68.0 | 0.0240 | |
| ICNet [34] | Fine, Validation | ImageNet | 69.5 | 0.0330 | - |
| ERFNet [26] | Fine | ImageNet | 69.7 | 0.0240 | - |
| DSNet-fast | Fine | ImageNet | 69.1 | 0.0189 | 0.0147 |

Table 7: The results of DSNet-fast and other methods on Cityscapes test set. The results of other methods are listed according to the online leaderboard and their reference papers (Cityscape webpage).

Furthermore, we compare DSNet-fast with the other state-of-the-art networks on the Cityscapes test set by submitting the test results to the online benchmark server. At the end, DSNet-fast achieves 69.1 % mIoU, as Table 7 shows. For the top-ranked methods, the architectures are more complex than DSNet-fast. Also, lots of data is included in their training procedure, resulting the better generalization and higher performance. Although the accuracy of DSNet is still lower than some high accuracy networks, DSNet is rather fast and accurate in competing with the efficient networks. It only takes 18.9 ms per image on a Titan X and 14.7 ms per image on a 1080Ti, for 1024×512 resolution inputs. The name of our method on the leaderboard is NCTU-ITRI.

## 5. Conclusions

In general, a deep learning model usually has high performance (accuracy) but often has a low inference speed. This makes the deep learning based methods difficult to apply into a real-world application. To solve the problem, we modify the network architecture based on the Fully Convolutional Network (FCN).

In order to find an efficient trade-off between accuracy and speed, we conduct a series of experiments. We explore a number of the encoder variations, examine the impact of input resolution, and the structure and the depth of a neural net. Next, we look into the fusion methods in the decoder, and ways to simplify the decoder. Finally, we propose an architecture that is able to process $1024 \times 512$ resolution images at 68 frames per second on a single 1080 Ti GPU card. In addition, our proposed architecture shows the good results on the two challenging road-scene datasets, CamVid and Cityscapes. This demonstrates that the proposed architecture is able to achieve a high speed and rather high accuracy processing.

### Acknowledgements

This work was supported in part by the Mechanical and Mechatronics Systems Research Lab., ITRI, Taiwan under Grant 3000547822. We would like to thank Shao-Yuan Lo for his helpful discussions during the course of this project.

# References

[1] V. Badrinarayanan, A. Kendall, and R. Cipolla, Segnet: A deep convolutional encoder-decoder architecture for image segmentation. *IEEE transactions on pattern analysis and machine intelligence,* 2017.

[2] G. J. Brostow, J. Fauqueur, and R. Cipolla, Semantic object classes in video: A high-definition ground truth database. *Pattern Recognition Letters,* 2009.

[3] L.-C. Chen, M. Collins, Y. Zhu, G. Papandreou, B. Zoph, F. Schroff, H. Adam, and J. Shlens, Searching for efficient multi-scale architectures for dense image prediction, in *NIPS*, 2018.

[4] L.-C. Chen, G. Papandreou, I. Kokkinos, K. Murphy, and A. L. Yuille, Deeplab: Semantic image segmentation with deep convolutional nets, atrous convolution, and fully connected crfs. *arXiv preprint arXiv:1606.00915,* 2016.

[5] L.-C. Chen, G. Papandreou, F. Schroff, and H. Adam, Rethinking atrous convolution for semantic image segmentation. *arXiv preprint arXiv:1706.05587,* 2017.

[6] L.-C. Chen, Y. Zhu, G. Papandreou, F. Schroff, and H. Adam, Encoder-decoder with atrous separable convolution for semantic image segmentation. *arXiv preprint arXiv:1802.02611,* 2018.

[7] M. Cordts, M. Omran, S. Ramos, T. Rehfeld, M. Enzweiler, R. Benenson, U. Franke, S. Roth, and B. Schiele, The cityscapes dataset for semantic urban scene understanding, in *CVPR*, 2016.

[8] J. Fu, J. Liu, Y. Wang, and H. Lu, Densely connected deconvolutional network for semantic segmentation, in *ICIP*, 2017.

[9] J. Fu, J. Liu, Y. Wang, and H. Lu, Stacked deconvolutional network for semantic segmentation. *arXiv preprint arXiv:1708.04943,* 2017.

[10] K. He, X. Zhang, S. Ren, and J. Sun, Deep residual learning for image recognition, in *CVPR*, 2016.

[11] K. He, X. Zhang, S. Ren, and J. Sun, Identity mappings in deep residual networks, in *ECCV*, 2016.

[12] G. Huang, Z. Liu, K. Q. Weinberger, and L. van der Maaten, Densely connected convolutional networks, in *CVPR*, 2017.

[13] S. Ioffe and C. Szegedy, Batch normalization: Accelerating deep network training by reducing internal covariate shift. *arXiv preprint arXiv:1502.03167,* 2015.

[14] M. A. Islam, M. Rochan, N. D. Bruce, and Y. Wang, Gated feedback refinement network for dense image labeling, in *CVPR*, 2017.

[15] S. Jégou, M. Drozdzal, D. Vazquez, A. Romero, and Y. Bengio, The one hundred layers tiramisu: Fully convolutional densenets for semantic segmentation, in *CVPR Workshop*, 2017.

[16] A. Kendall, V. Badrinarayanan, and R. Cipolla, Bayesian segnet: Model uncertainty in deep convolutional encoder-decoder architectures for scene understanding. *arXiv preprint arXiv:1511.02680,* 2015.

[17] T.-Y. Lin, M. Maire, S. Belongie, J. Hays, P. Perona, D. Ramanan, P. Dollár, and C. L. Zitnick, Microsoft coco: Common objects in context, in *ECCV*, 2014.

[18] S.-Y. Lo, H.-M. Hang, S.-W. Chan, and J.-J. Lin, Efficient dense modules of asymmetric convolution for real-time semantic segmentation. *arXiv preprint arXiv:1809.06323,* 2018.

[19] J. Long, E. Shelhamer, and T. Darrell, Fully convolutional networks for semantic segmentation, in *CVPR*, 2015.

[20] S. Mehta, M. Rastegari, A. Caspi, L. Shapiro, and H. Hajishirzi, ESPNet: Efficient Spatial Pyramid of Dilated Convolutions for Semantic Segmentation. *arXiv preprint arXiv:1803.06815,* 2018.

[21] S. Mehta, M. Rastegari, L. Shapiro, and H. Hajishirzi, ESPNetv2: A Light-weight, Power Efficient, and General Purpose Convolutional Neural Network. *arXiv preprint arXiv:1811.11431,* 2018.

[22] H. Noh, S. Hong, and B. Han, Learning deconvolution network for semantic segmentation, in *ICCV*, 2015.

[23] A. Paszke, A. Chaurasia, S. Kim, and E. Culurciello, Enet: A deep neural network architecture for real-time semantic segmentation. *arXiv preprint arXiv:1606.02147,* 2016.

[24] A. Paszke, S. Gross, S. Chintala, G. Chanan, E. Yang, Z. DeVito, Z. Lin, A. Desmaison, L. Antiga, and A. Lerer, Automatic differentiation in PyTorch. 2017.

[25] R. P. Poudel, U. Bonde, S. Liwicki, and C. Zach, Contextnet: Exploring context and detail for semantic segmentation in real-time. *arXiv preprint arXiv:1805.04554,* 2018.

[26] E. Romera, J. M. Alvarez, L. M. Bergasa, and R. Arroyo, Erfnet: Efficient residual factorized convnet for real-time semantic segmentation. *IEEE Transactions on Intelligent Transportation Systems,* 2018.

[27] O. Russakovsky, J. Deng, H. Su, J. Krause, S. Satheesh, S. Ma, Z. Huang, A. Karpathy, A. Khosla, and M. Bernstein, Imagenet large scale visual recognition challenge. *International Journal of Computer Vision,* 2015.

[28] K. Simonyan and A. Zisserman, Very deep convolutional networks for large-scale image recognition. *arXiv preprint arXiv:1409.1556,* 2014.

[29] N. Srivastava, G. Hinton, A. Krizhevsky, I. Sutskever, and R. Salakhutdinov, Dropout: A simple way to prevent neural networks from overfitting. *The Journal of Machine Learning Research,* 2014.

[30] C. Szegedy, V. Vanhoucke, S. Ioffe, J. Shlens, and Z. Wojna, Rethinking the inception architecture for computer vision, in *CVPR*, 2016.

[31] M. Treml, J. Arjona-Medina, T. Unterthiner, R. Durgesh, F. Friedmann, P. Schuberth, A. Mayr, M. Heusel, M. Hofmarcher, and M. Widrich, Speeding up semantic segmentation for autonomous driving, in *NIPS Workshop*, 2016.

[32] A. Valada, R. Mohan, and W. Burgard, Self-Supervised Model Adaptation for Multimodal Semantic Segmentation. *arXiv preprint arXiv:1808.03833,* 2018.

[33] F. Yu and V. Koltun, Multi-scale context aggregation by dilated convolutions. *arXiv preprint arXiv:1511.07122,* 2015.

[34] H. Zhao, X. Qi, X. Shen, J. Shi, and J. Jia, Icnet for real-time semantic segmentation on high-resolution images, in *ECCV*, 2018.

[35] H. Zhao, J. Shi, X. Qi, X. Wang, and J. Jia, Pyramid scene parsing network, in *CVPR*, 2017.

[36] Y. Zhuang, L. Tao, F. Yang, C. Ma, Z. Zhang, H. Jia, and X. Xie, RelationNet: Learning deep-aligned representation for semantic image segmentation, in *ICPR*, 2018.

[37] Y. Zhuang, F. Yang, L. Tao, C. Ma, Z. Zhang, Y. Li, H. Jia, X. Xie, and W. Gao, Dense Relation Network: Learning Consistent and Context-Aware Representation for Semantic Image Segmentation, in *ICIP*, 2018.